# Computer vision-enriched discrete choice models, with an application to residential location choice

Sander van Cranenburgh[1]
Francisco Garrido-Valenzuela[1]
[1]CityAI lab, Transport and Logistics Group, Delft University of Technology

Keywords
Discrete choice modelling, Computer vision, Residential location choice, Commute, Travel behaviour

*Abstract*
*Visual imagery is indispensable to many multi-attribute decision situations. Examples of such decision situations in travel behaviour research include residential location choices, vehicle choices, tourist destination choices, and various safety-related choices. However, current discrete choice models cannot handle image data and thus cannot incorporate information embedded in images into their representations of choice behaviour. This gap between discrete choice models' capabilities and the real-world behaviour it seeks to model leads to incomplete and, possibly, misleading outcomes. To solve this gap, this study proposes "Computer Vision-enriched Discrete Choice Models" (CV-DCMs). CV-DCMs can handle choice tasks involving numeric attributes and images by integrating computer vision and traditional discrete choice models. Moreover, because CV-DCMs are grounded in random utility maximisation principles, they maintain the solid behavioural foundation of traditional discrete choice models. We demonstrate the proposed CV-DCM by applying it to data obtained through a novel stated choice experiment involving residential location choices. In this experiment, respondents faced choice tasks with trade-offs between commute time, monthly housing cost and street-level conditions, presented using images. As such, this research contributes to the growing body of literature in the travel behaviour field that seeks to integrate discrete choice modelling and machine learning.*

## 1. Introduction

Discrete Choice Models (DCMs) are widely used in transportation (and beyond) to describe how individual choices result from preferences over attributes and available alternatives in multi-attribute decision-making. When DCMs were incepted in the 1970s, they were used to explain and predict mode and destination shares (McFadden 1974; McFadden 2001). Nowadays, DCMs are applied to a wide variety of choice situations, including residential location choice, route choice, vehicle choice, airport choice, time of day choice and many more (de Jong et al. 2003; Guevara and Ben-Akiva 2006; Hess et al. 2007; Prato 2009; Pinjari et al. 2011; Beck et al. 2013; Hess and Daly 2014). DCMs are built on the notion that attributes have numeric values or can be converted into numeric values, e.g. in the case of a categorical level. In other words, the attributes that jointly make an alternative only involve numbers.

Visual imagery is crucial to many multi-attribute decision situations, in and beyond transportation. For example, visual information is indispensable to residential location choices. In today's digital age, it is hard to imagine searching for a house on a real estate website without access to images. Other examples of such decision situations in transportation include vehicle choices, tourist destination choices, transport infrastructure design choices and choices related to safety, such as where to cross a street on foot and whether a route is safe enough to cycle. The widespread use of visual imagery, e.g. on websites like Zillow.com and in Stated Choice (SC) experiments, can be attributed to the fact that it is easier for people to perceive and process information presented through images than information presented in text or numbers (Pinker 1990). In addition, visual imagery provides valuable details about the alternative, such as scale, texture, or quality, that are difficult to convey through textual descriptions or numbers (Childers et al. 1985). For instance, in a residential location choice context, street-level conditions such as "safeness", "openness", and "greenness" cannot be easily expressed in numbers but can effectively be communicated by images. The COVID-19 pandemic brought the importance of street-level conditions to the forefront, with millions of white-collar workers relocating to suburban areas with better street-level conditions during the pandemic-induced remote work shift (Economist 2022; Lee and Huang 2022). Therefore, to accurately represent choice behaviour in multi-attribute situations that involve visual imagery, it is necessary to have choice models capable of working with image data.

However, present-day DCMs cannot handle image data and, therefore, cannot incorporate information from images into their representations of choice behaviour. The inability to handle image data in DCMs creates a stark contrast between the behaviour it seeks to model, where images are widely used, and what DCMs can do. Even when researchers deliberately use images in SC experiments to visualise information that is challenging to convey in numbers, the information embedded in the images is scantly accounted for (Cherchi and Hensher 2015; see Hevia-Koch and Ladenburg 2019 for a thorough discussion). DCMs' inability to handle image data leads to incomplete and potentially misleading outcomes.

As a solution, this study proposes "Computer Vision-enriched Discrete Choice Models" (henceforth abbreviated as CV-DCMs). These models can handle choice tasks involving both numeric attributes and an image. CV-DCMs are grounded in Random Utility Maximisation (RUM) principles (McFadden 2000; Hess et al. 2018). Therefore, CV-DCMs maintain the solid behavioural foundation of traditional DCMs while expanding their application to include image data. We demonstrate the effectiveness of the proposed CV-DCMs by shedding light on the importance of street-level conditions to residential location choice behaviour relative to travel-related factors, such as travel time and travel cost. To do so, we have developed and administered a novel stated choice experiment involving trade-offs between commute travel time, monthly housing cost (both numeric attributes) and street-level conditions (image).

Methodologically, this study contributes to the growing body of literature in the travel behaviour field that seeks to integrate machine learning and DCMs (e.g. Iglesias et al. 2013; Hurtubia et al. 2015; Rossetti et al. 2019; Sifringer et al. 2020; Arkoudi et al. 2021; Ramírez et al. 2021; van Cranenburgh et al. 2021; Szép et al. 2023). Our model shares similarities with Arkoudi et al. (2021) in its use of embeddings, but it extends this work by incorporating images. We also build on Ramírez et al. (2021) approach, which also seeks to explain choice behaviour in the presence of images. But, importantly, our model does not rely on object detection and segmentation models, which can detect only a finite and predefined number of objects. Also, this work contributes to the growing body of literature in travel behaviour research using computer vision techniques more generally (e.g. Yencha 2019; Ito and Biljecki 2021; Basu and Sevtsuk 2022). Furthermore, substantively, this research contributes to the literature on residential location choice behaviour by highlighting the importance of street-level conditions in residential location choices.

The remaining part of this paper is organised as follows. Section 2 describes the proposed CV-DCMs. Section 3 discusses the stated choice data collection effort and reports the sample statistics, descriptive results and details on the training of the model. Section 4 contains the main results. Section 4.1 presents the results from the CV-DCMs and compares model fit and parameter estimates with those of traditional discrete choice models, which do not account for images. Section 4.2 shows what the CV-DCM has learned about what decision-makers find relevant for their residential location choices. It provides face validity to the modelling results. Section 4.3 demonstrates the merits of the CV-DCM by showing how CV-DCMs can be used to deepen understanding of residential location preferences.

## 2. Methodology

This section presents the methodology. Section 2.1 introduces relevant models and concepts from computer vision. Section 2.2 proposes the modelling framework. Section 2.3 briefly discusses implementation details and training.

### 2.1. Preliminary: computer vision models and concepts

Computer Vision (CV) is concerned with extracting meaningful information from images, videos, and other forms of visual data. CV models typically detect scenes and objects in images (Gu et al. 2018). Nowadays, CV models are applied in a wide range of applications and numerous fields. In transport, CV models are essential for future autonomous vehicles to perceive and understand their environment; in healthcare, CV models are used in medical imaging to aid in diagnosing diseases and abnormalities; and, in retail, CV models are used to track customer movement in stores. As CV models grow and become more powerful, they can perform increasingly sophisticated visual tasks (Sevilla et al. 2022). The largest CV models currently in use contain over 1 billion weights (Zhai et al. 2022).

The building blocks of images are pixels. A pixel represents a single point in an image and contains information about its colour and brightness. Each pixel has a spatial location ($h$ x $w$) and a colour value. Most colour images nowadays use three colour channels: Red (R), Green (G), Blue (B), and 8 bits per colour channel (implying three 0-225 values), with which it is possible to create a wide range of colours and shades. Mathematically, images are usually represented as 3D tensors, which are multi-dimensional arrays of numerical values. Tensors enable easy processing and manipulation of images using various mathematical operations and algorithms. The three dimensions of an image tensor typically correspond to the image's width, height, and colour channels. Thus, an RGB colour image with a resolution of 900 x 600 pixels can be represented as a 3D tensor with a shape of (900, 600, 3), where the first two dimensions correspond to the height and width of the image and the third dimension corresponds to the colour channels. An image tensor of a 900 x 600 RGB colour image contains 1.6m data points.

CV models typically have two main components: a feature extractor and a classifier. The feature extractor generally is a deep neural network which is trained to extract relevant features from images. The output of the feature extractor is the feature map, which is a lower-dimensional vector representation of the image and captures its salient features. In other words, the feature map contains (most of) the information of the image but is more compact in form. Usually, a feature map (a.k.a. embedding) is a flat array of floating points. The classifier is a separate component which is trained to classify the input image based on the feature map extracted by the feature extractor. Typically, the classifier is a Multilayer Perceptron (MLP) with one or more fully connected layers, producing a probability distribution over the different output classes.

Traditionally, computer vision models have used Convolution Neural Networks (CNNs) architectures; see Figure 1. A CNN comprises a series of so-called convolution layers, which reduce the dimensionality of the data by taking a series of convolutions followed by an MLP classifier. More recently, transformer-based architectures have taken over the CV field (Dosovitskiy et al. 2020). In contrast to a CNN-based architecture, so-called Vision Transformers (ViT) rely on self-attention and multi-head attention mechanisms to learn spatial relationships between different parts of the image. In a ViT architecture, the input image is divided into a grid of non-overlapping patches, which are linearly embedded to produce a sequence of feature maps. These feature maps are then processed by a series of transformer encoder layers to learn spatial relationships between the different parts of the image. Finally, an MLP classifier is applied to the feature map in the last layer to produce the final classification result.

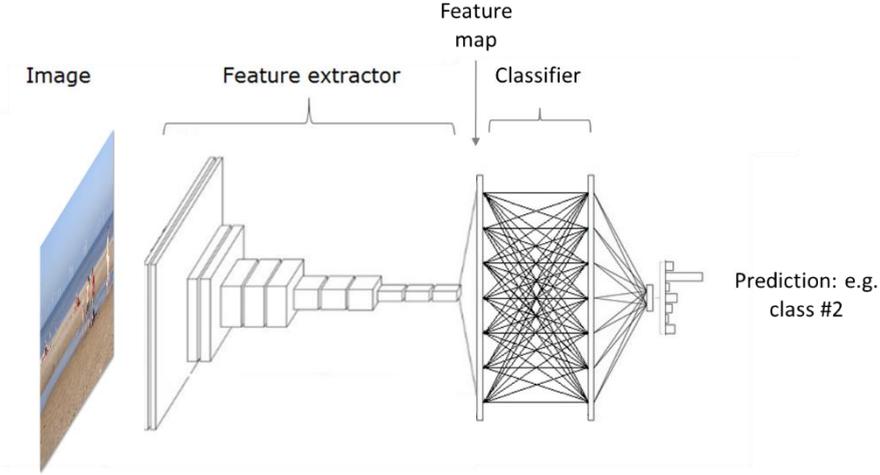

Figure 1: CNN-based architecture

## 2.2. Computer vision-enriched discrete choice models

Throughout this paper, we consider the following choice situation. A decision-maker, *n*, faces a multi-attribute choice task with a set of *J* mutually exclusive alternatives. Each alternative is described by *M* numeric attributes $X_j = \{x_{j1}, x_{j2}, \ldots, x_{jM}\}$, such as e.g. travel cost and travel time and by a (colour) image $S_j$ with a resolution of $H \times W \times C$.

We assume decision-makers make decisions based on Random Utility Maximising (RUM) principles (McFadden 1974), see Equation 1, where $U_{jn}$ denotes the total utility experienced by decision-maker *n* considering alternative *j*, $V_{jn}$ is the utility experienced by decision-maker *n* derived from attributes observable by the analyst. And, to account for the fact that the analyst does not observe everything that matters to the decision-maker's utility, an additive error term $\varepsilon_{jn}$ is added to each alternative (Train 2003).

$$U_{jn} = V_{jn} + \varepsilon_{jn} \qquad \text{Equation 1}$$

Furthermore, we assume decision-makers experience utility from both the numeric attributes $X_j$ and image $S_j$, see Equation 2, where *v* is a preference function which maps the attributes and image of an alternative onto utility.

$$U_{jn}(X_{jn}, S_{jn}) = v(X_{jn}, S_{jn}) + \varepsilon_{jn} \qquad \text{Equation 2}$$

In addition, we make three more assumptions to develop the CV-DCM:

1. We assume that the utility derived from the numeric attributes and the image are separable and additive in utility space, see Equation 3, where function *f* maps the (observed) numeric attributes onto utility and function *g* maps the relevant information from the image onto utility.

$$U_{jn}(X_{jn}, S_{jn}) = f(X_{jn}) + g(S_{jn}) + \varepsilon_{jn} \qquad \text{Equation 3}$$

2. We assume that utility is linear and additive with numeric attributes as well as with images' feature maps. Thus, $f$ and $g$ are standard linear-additive utility functions. As discussed in section 2.1, feature maps are more compact representations of images. Accordingly, we let $Z_j = \{z_{j1}, z_{j2}, ..., z_{jK}\}$ denote the feature map of image $S_j$, and $\varphi(w): \mathbb{R}^{H \times W \times C} \to \mathbb{R}^K$ be a function that maps image $S_j$ onto feature map $Z_j$. Hence, $\varphi$ is the transformation produced by the feature extractor of a CV model, and $w$ are its associated weights (i.e., parameters). Both the numeric attributes $X_j$ and feature map $Z_j$ enter the utility function in a linear-additive fashion, as shown in Equation 4. In Equation 4, $\beta_m$ denotes the marginal utility associated with attribute $m$; $x_{jmn}$ denotes the attribute level of numeric attribute $m$ of alternative $j$, as faced by decision-maker $n$; and $\beta_k$ denotes the weight associated with the $k^{\text{th}}$ element of feature map $Z_{jn}$.

$$U_{jn} = \underbrace{\sum_m \beta_m x_{jmn}}_{\substack{\text{Utility derived} \\ \text{from numeric attributes}}} + \underbrace{\sum_k \beta_k z_{jkn}}_{\substack{\text{Utility derived} \\ \text{from image feature map}}} + \varepsilon_{jn} \qquad \text{where } Z_{jn} = \varphi(S_{jn} \mid w) \qquad \text{Equation 4}$$

The reason that we let feature maps, as opposed to individual pixel values, enter the utility function is twofold. First, the pixel values in and of themselves are behaviourally meaningless. Rather, the joint distribution of many pixels together produces higher-level behaviourally meaningful concepts that can be expected to generate utility. A feature map can be trained to provide a representation that can be mapped linearly (Goodfellow et al. 2016 pg. 518). Second, a medium-sized colour image of 640 x 480 pixels contains almost 1 million data points. This high dimensionality renders using each pixel directly in the utility function practically infeasible. Instead, $\varphi(w)$ is learned to produce a feature map containing the relevant information of the image to the choice behaviour in a way that maps linearly onto utility.

3. In line with common practice in choice modelling, we assume $\varepsilon_{jn}$ is independent and identically Extreme Value Type I distributed with a variance of $\pi^2/6$, resulting in the well-known and convenient closed-form logit formula for the choice probabilities ($P_{in}$), given in Equation 5, where $C_n$ denotes the set of alternatives presented to decision maker $n$. Note that this assumption would, from a machine learning perspective, be equivalent to saying that the output layer is a Softmax function.

$$P_{in} = \frac{e^{V_{in}}}{\sum_{j \in C_n} e^{V_{jn}}} \qquad \text{Equation 5}$$

Figure 2 depicts a graphical representation of the model structure of the proposed CV-DCM model. An essential aspect of the CV-DCM model's architecture is that the network's upper (associated with the left-hand side alternative) and lower (associated with the right-hand side alternative) parts are identical. Thereby, the left and right-hand side alternatives are mathematically treated in the same way, allowing us to interpret the values on the nodes of the last layer as a utility. In the computer vision field, this network structure (bar the numeric attributes) is called a Siamese network (Bromley et al. 1994). Siamese networks are typically used to determine the (dis)similarity between two images. Importantly, even though we can interpret the last layer as utility, we cannot interpret $β_k$ in the same way we can with $β_m$. $β_k$ can be conceived as a marginal utility –after all, it reflects the change in utility by a unit change in the attribute level. But, because the meaning and units of the elements on the feature map, $Z_j$, are unclear, they do not carry a behavioural meaning.

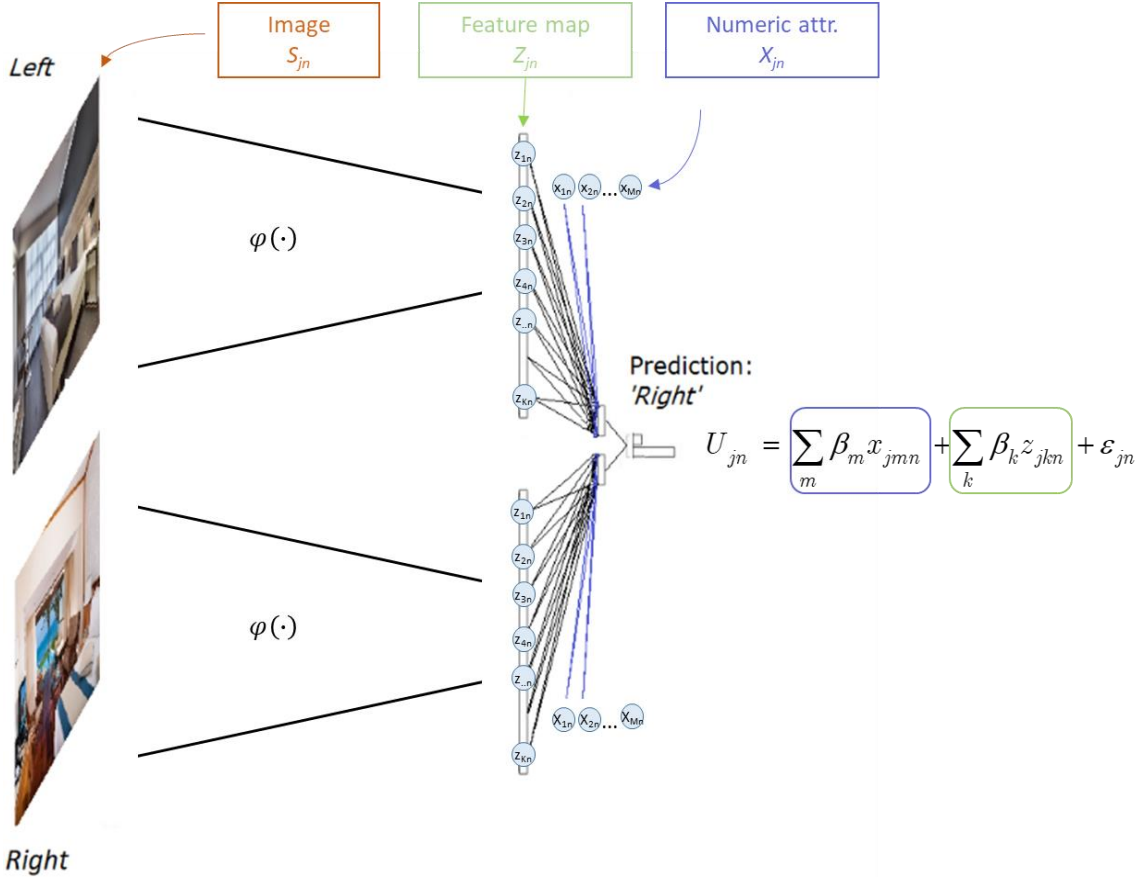

Figure 2: Model structure of CV-DCM

### 2.3. Feature extractor and training

In this study, we use the feature extractor of the DeiT base model (Touvron et al. 2021). DeiT models are data-efficient vision transformer-based models which produce competitive capabilities on benchmark data sets, such as ImageNet (Russakovsky et al. 2015), at a lower computational cost and data requirements than many of its competitors (Touvron et al. 2021). The DeiT base model consumes a relatively modest 86 million weights. Furthermore, we use transfer learning to train our CV-DCM (Bengio 2012) to lower the computational time and amount of training data. The idea of transfer learning is to use a pre-trained network as the starting point for developing another network for a closely related task. In other words, rather than retraining the whole model from scratch, we start the training from an already good starting point when we train the CV-DCM. Our pre-trained DeiT base model is trained on ImageNet (Deng et al. 2009); a widely used benchmark image data set containing 1.2 million training images with 1,000 object classes.

### 3. Data collection and training

We demonstrate the proposed CV-DCM by applying it to data obtained through a stated choice experiment involving residential location choices. The residential location choice makes a suitable case study because both numeric attributes and visual information can be expected to be important to residential location choice behaviour (Smith and Olaru 2013). Moreover, images of the sort that we need for modelling residential location choice behaviour, namely street-view images, are widely available from map services such as Google, Apple, and Baidu and have been used in numerous scientific inquiries, including research on safety perceptions and people's density in urban places (Dubey et al. 2016; Ito and Biljecki 2021; Ma et al. 2021; Garrido-Valenzuela et al. 2022). Having access to a sufficiently large and diverse set of images is crucial for effectively training the feature extractor of the CV-DCM. While the exact number of images required is unknown before training, more images (and choice observations) generally lead to better training. In addition to their availability, street-view images have been shown to be a reliable representation of street-level conditions, as demonstrated by (Hanibuchi et al. 2019).

### 3.1. Stated choice experiment

In the Stated Choice (SC) experiment, we asked respondents to imagine they were required to move to a different neighbourhood. They were presented with two alternatives for residential locations and asked to indicate which of the two they would choose. Figure 3 shows a screenshot of a choice task from the experiment. Prior to starting the choice experiment, respondents were provided with the following information:

1. Your new house is identical to your current house in terms of, e.g. size, type, built-year, furniture, maintenance, etc. Only your neighbourhood changes.
2. Your monthly housing cost (including rent, mortgage, taxes, insurance, etc.) may go up or go down.
3. Your new neighbourhood is relatively near your current neighbourhood, but your commute time may still go up or down. The commute time is for your current mode of transport.
4. Your situation stays the same in all other aspects, e.g. in terms of distances to amenities, schools, the general practitioner, etc.
5. The images shown in the choice tasks depict the window view at ground level on the street side.

The alternatives comprise two salient numeric attributes: monthly housing costs (*hhc*) and commute travel time (*tti*). We choose these two attributes for three reasons. Firstly, they are known to be important to the residential location choice (Tillema et al. 2010). Secondly, they apply generically to almost everyones' residential location choice. Thirdly, they may help to interpret our empirical results. The combination of cost and time attributes allows us to compute the Value-of-Travel-Time (VTT), a metric that is widely studied in transport (Small 2012) and thus can be used for model validation. Finally, we did not include more attributes to the design because the paper's objective is to demonstrate the effectiveness of the proposed CV-DCMs to capture visual preferences instead of, e.g. developing a comprehensive model to predict residential location choices.

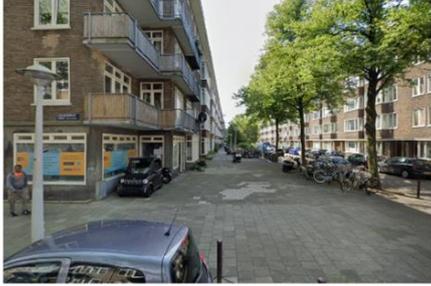

Figure 3: Screenshot of the pivoted stated choice experiment
(translated to English; original in Dutch)

As can be seen in Figure 3, we have opted for a pivoted experimental design. We use a pivoted design to present respondents with as realistic choice situations as possible. Using absolute levels instead of pivoted levels would presumably render many choice tasks unrealistic because of the considerable variation across respondents' current situations, especially regarding housing costs. For the attribute housing cost, we have used seven pivoted levels. For the attribute travel time, the number of levels and ranges we presented to the respondent depended on the respondent's current travel time, see Table 1. The ranges of both attributes were determined through a small pilot conducted before the actual survey.

Table 1: Attribute levels Stated Choice experiment

| Current commute travel time of the respondent ($TT_n$) | Attribute levels | |
|---|---|---|
| | Housing cost (*hhc*) [€] | Commute travel time (*tti*) [minutes] |
| $TT_n$ < 10 minutes | N/A | |
| 10 minutes < $TT_n$ < 20 minutes | -225, -150, -75, 0, +75, +150, +225 | -5, 0, +5, +10, +15 |
| 20 minutes < $TT_n$ < 30 minutes | | -10, -5, 0, +5, +10, +15 |
| 30 minutes < $TT_n$ | | -15, -10, -5, 0, +5, +10, +15 |

### 3.1.1. Street-view images

Besides monthly housing costs and commute travel time, each alternative comes with an image showing the street-level conditions. This image is randomly sampled from a database of street-view images we created before conducting the stated choice experiment. A major effort went into the construction of this database with street-view images. Specifically, we took the following steps to build the database. First, we randomly selected 50 municipalities (of about 350) in the Netherlands. We capped the number of municipalities to 50 because using more would lead to collecting many more images than we would need for our SC experiment. Second, we created a grid of points with 150-metre spacing within areas designated as residential areas (within the selected municipalities). Third, we retrieved the nearest street-view image id for each point on the grid using Google's API. We collected ids for all available images taken in 2020, or later. Each image id corresponds to a 360-degree panorama photo. Fourth, from each panorama, we generated two image urls with 90-degree angles to the direction of the street (to both directions). This latter ensures the images are 'window views' (e.g. as opposed to views parallel to the driving direction of the Google car). Finally, urls of images of poor quality were algorithmically removed. More specifically, urls to black images, blurred images and images with tilted horizons were removed. The final database contains the urls a little over 60k street-view images of residential streets from 50 municipalities in the Netherlands.

Importantly, for each image in our database, we also stored the month of the year in which the image was taken. The Netherlands lies in temperate zones, having four distinct seasons. Even though Google collects images on dry days only, due to the seasonality, street-view images taken in the winter may look different from those taken in summer. These differences might, in turn, impact the utility experienced by the respondent from the depicted local environment (and thus must be accounted for in our models).

### 3.1.2. Experimental design

We have used a random experimental design. Because the images do not possess ordinal or categorical levels, adopting an orthogonal or efficient experimental design strategy was not feasible, at least not considering the images. Therefore, we took a two-step approach to construct the choice tasks. First, we randomly pulled a pair of images from our image database. The only requirement imposed on the drawing was that the drawn images were not from the municipality where the respondent lives. We determined each respondent's municipality (and province) based on the postcode we elicited at the start of the survey. We excluded images from the respondent's municipality to avoid unobserved heterogeneity entering our experiment derived from respondents' knowledge of places the images were taken. Unobserved utilities flowing into stated choice experiments could lead to biased modelling outcomes if not econometrically accounted for (see e.g., Train and Wilson 2008; Van Cranenburgh et al. 2014; Guevara and Hess 2019). While excluding images from respondents' own municipalities does not guarantee that respondents do not recognise the places the street-view images were taken, it lowers the probability.

Second, we added the housing cost (*hhc*) and travel time (*tti*) levels. To do so, we randomly pulled a choice task from one of three tables with choice tasks we generated before conducting the SC experiment. Each table was created by taking the following steps. First, a full-factorial design was created based on the attribute levels shown in Table 1. Second, we excluded choice tasks that did not involve a trade-off between housing costs and travel time. Removing such (partially) dominating choice tasks is possible because we have strong prior beliefs for the expected sign of the preference parameters for housing cost and travel time. Third, we excluded all choice tasks where one or more attribute levels were equal. As a result of this choice task construction approach, each choice task necessarily consists of a trade-off between housing cost and travel time.

### 3.2. Data collection and sample description

The survey was implemented in SurveyEngine software and conducted in September 2022. The survey started with a few questions to determine respondents' eligibility for the survey. In particular, we elicited respondents' age, gender, postcode, and current commute travel time. Then came the SC experiment, in which each respondent was presented with 15 choice tasks. The images used in the choice tasks were directly retrieved from Google servers based on the urls from our image database. The survey ended with a series of questions regarding the respondents' current housing situation (e.g. housing costs, rating of the current visual street-level conditions) and commute situation (e.g. mode of transport, number of commute days). Noteworthy, we also asked respondents how important the three attributes (housing cost, travel time and the image) were for their decisions on a scale from 1 to 10. Although it is well-known that direct elicitation of preferences is treacherous (Nisbett and Wilson 1977), it still can provide first (albeit inconclusive) evidence of the importance of the presented street-view images relative to the numeric attributes for the residential location choices.

The target population for the survey was the Dutch population of 18 years and older, with ten or more minutes of commute travel time. The latter requirement was necessary because we used a pivoted experimental design. Because of this latter condition, no official population statistics exist to compare our sample against, but we do not expect this condition to affect the population statistics substantially. Therefore, care was taken that the sample was, by and large, representative of the Dutch 18 years and older population in terms of gender, age and spatial distribution across the Netherlands. Cint[1], a panel data provider, provided the panel of respondents. In total, 800 respondents completed our survey.

Table 2 shows the sample statistics. Overall, the sample is representative of the target population. Also, for the variables that are not explicitly considered during the data collection, such as e.g. the modal split and household composition, the statistics are close to the population data (c.f. Ton et al. 2019). Furthermore, looking at the reported monthly housing cost, we notice that the largest share of the respondents has a housing cost below €750. This seems reasonable since the average net housing cost of rental houses in the Netherlands is around €700 p/m; homeowners' average net housing cost is slightly above €900 p/m (Stuart-Fox et al. 2022).

Table 2: Sample statistics

| Socio-demographic variable | Category | Distribution |
| --- | --- | --- |
| Age | 18 - 29 year | 21% |
| | 30 - 39 year | 19% |
| | 40 - 49 year | 20% |
| | 50 - 59 year | 22% |
| | 60 - 69 year | 17% |
| | +70 year | 1% |
| Gender | Male | 50% |
| | Female | 50% |
| Province | North (Groningen, Friesland, Drenthe) 1,3,5 | 12% |
| | East (Gelderland, Overijssel) | 23% |
| | South (Limburg, Noord-Brabant, Zeeland) | 24% |
| | West (N-Holland, Z-Holland, Utrecht, Flevoland) | 41% |
| Current commute travel time (TT) | 10 minutes < TT < 20 minutes | 35% |
| | 20 minutes < TT < 30 minutes | 31% |
| | 30 minutes < TT < 45 minutes | 20% |
| | 45 minutes < TT | 14% |
| Primary mode for commute | Bike, E-bike, Scooter, Moped | 30% |
| | Bus, Metro, Tram | 8% |
| | Train | 10% |
| | Car, Motor bike | 52% |

---

[1] See www.cint.com

| | | |
|---|---|---|
| Commuting days per week | 1 day per week | 8% |
| | 2 days per week | 15% |
| | 3 days per week | 20% |
| | 4 days per week | 22% |
| | 5 or more days per week | 35% |
| Household composition | One-person household | 26% |
| | Multiple-person household without children | 40% |
| | Multiple-person household with children | 34% |
| House type | Flat, gallery, porch, apartment | 23% |
| | Terraced house | 31% |
| | Corner house | 16% |
| | Semidetached house | 14% |
| | Detached house | 15% |
| Current monthly housing cost (HC) | HC < 750 p/m | 36% |
| | 750 p/m < HC < 1,250 p/m | 33% |
| | 1,250 p/m < HC < 1,750 p/m | 16% |
| | 1,750 p/m < HC | 6% |
| | I do not want to report | 9% |
| Rating of own visual street-level conditions | 1 (worst) | 1% |
| | 2 | 6% |
| | 3 | 21% |
| | 4 | 46% |
| | 5 (best) | 26% |

### 3.3. Descriptive analysis

Figure 4 shows histograms of the self-reported importance levels of the street-view images (left), monthly housing costs (middle) and commute travel times (right). Figure 4 shows that the street-view images and monthly housing costs are, on average, considered equally important to the residential location choice and more important than commute travel times. The variance in the ratings across respondents is higher for the street-view images than for the monthly housing cost – suggesting a considerable amount of preference heterogeneity is present in the importance of street-level conditions. However, we observe the most variance for the commute travel time. Noteworthily, the importance rating for the street-view images is weakly negatively correlated with the importance ratings for monthly housing costs ($\rho = -0.10$) and uncorrelated with the ratings for commute travel time ($\rho = 0.02$). In contrast, the importance ratings for monthly housing costs and commute travel times are strongly positively correlated ($\rho = 0.36$). This strong positive correlation reveals that people who find housing costs important usually also find commute travel time important, and vice versa.

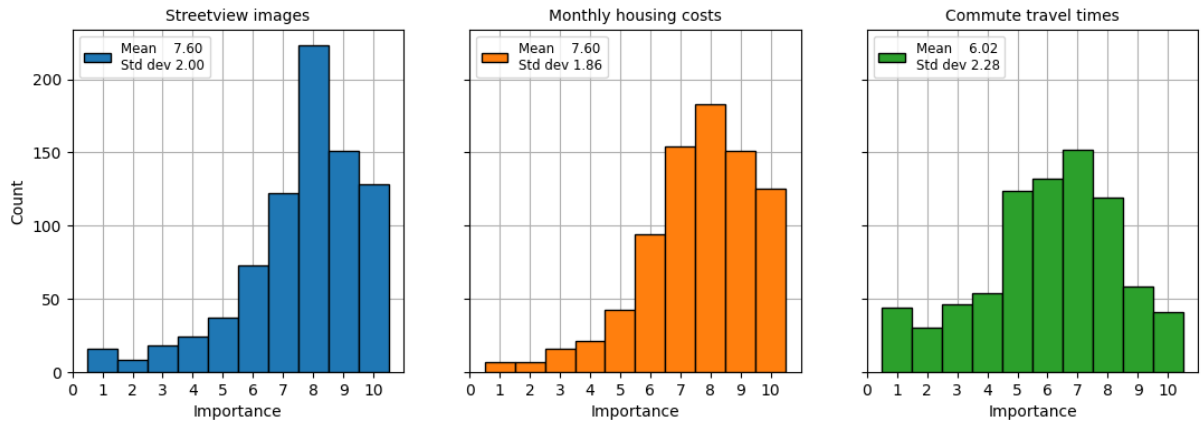

Figure 4: Self-reported importance levels of attributes in the SC experiment

Figure 5 shows the Pearson correlation coefficients between importance ratings and a selection of respondent characteristics. Interestingly, the top row shows that the importance of the street-view images correlates strongest with the self-reported rating of respondents' current visual street-level conditions. This strong positive correlation suggests that people living in visually attractive neighbourhoods consider their visual street-level conditions relatively more important than people living in visually less attractive places. This observation aligns with Lee and Waddell (2010), who also find that the current situation affects residential location choice behaviour. Moreover, we see that the importance of the street-view images positively correlates with living in a detached or semi-detached house. A self-selection mechanism could explain this effect: people caring about their visual street-level conditions are more likely to choose an attractive residential location (see e.g., Van Wee 2009 for discussions on self-selection effects in residential location choices; Cao 2014). Finally, perhaps somewhat counter to expectations, we see that variables such as gender and monthly housing costs do not strongly correlate with the importance given to the street-view images.

Furthermore, Figure 5 reveals that the importance of the monthly housing cost (middle row) correlates strongest with living in house type 'Flat, gallery, porch, or apartment'. This correlation seems in line with intuition, given that low-income people are more likely to live in this type of housing. Finally, we see that the importance of the commute travel time (bottom row) positively correlates with age class 18-39 years. Since this age class sits in the centre of the working-age population, it makes sense that commute travel time is essential to this group. Altogether, the correlations reported in Figure 5 seem plausible.

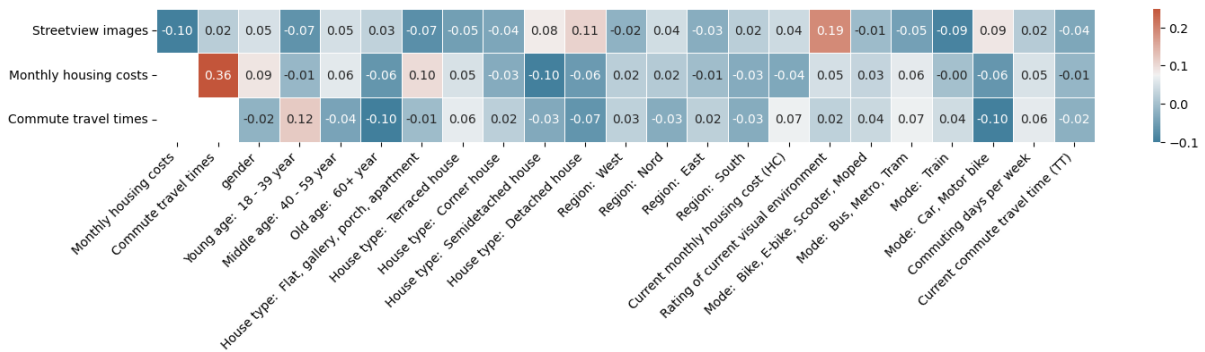

Figure 5: Pearson correlation coefficients between importance ratings and respondent characteristics

Next, we analyse the images used in the stated choice experiment. Although our street-view image database comprises urls to over 60k images, only slightly over 7.5k unique images are used in the stated choice experiment. Because images are drawn randomly from our image database with replacement, we expect that some images will be sampled more than once. Indeed, most images are used once. However, counter to our design intentions, some images are used 20 times or more. A possible underlying cause could be the seed numbers used by the survey platform's software. Nevertheless, regardless of this issue's origin, when we deal with the issue carefully during the training of our models (see Section 3.4.4), it will not have to impact our (substantive) findings.

Finally, Figure 6 shows the distribution of the month of the year of the images used in the survey. In line with expectations, the images are not evenly distributed over the year. We see that most images are taken in spring and summer (March to September). Furthermore, we notice that images are sampled for all 12 months. This implies we can account for the impact of the seasons on the utility derived from the street-view images by estimating constants for all months (except one, which we need to fix to zero for normalisation).

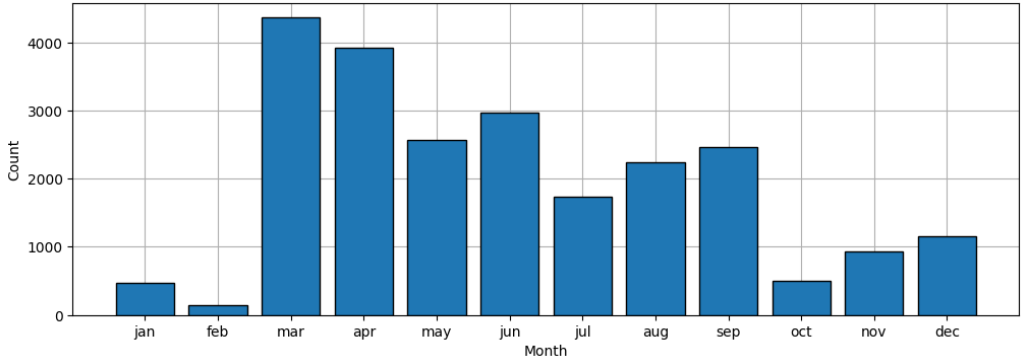

Figure 6: Distribution of images used in the stated choice experiment over the months of the year

### 3.4. Training
### 3.4.1. Loss function and implementation
Training a CV-DCM involves finding the weights of the model ($\beta$, $w$) that minimise the loss function. In other words, the weights of the feature extractor and preference parameters of the utility function are jointly optimised. For this study, we use a cross-entropy loss function with an L2 regularisation term, see Equation 6. Minimising the cross-entropy loss is equivalent to maximising the Log-Likelhood (*LL*) of the data given the model – which is common practice in the choice modelling literature. The L2 regularisation aims to reduce the chance of model overfitting by penalising the magnitude of the weights in the model. $\gamma$ governs the strength of the regularisation. Note that we apply regularisation only to $w$ and not to preference parameters $\beta_m$ and $\beta_k$. Regularising preference parameters could lead to undesirable biases.

$$w^*, \beta^* = \arg\min_{w,\beta} \left[ \overbrace{\frac{1}{N} \sum_{n=1}^{N} \sum_{j=1}^{J} y_{nj} \log\left(P_{nj} \mid X_{nj}, S_{jn}, \beta\right)}^{\text{cross-entropy loss}} + \overbrace{\gamma \sum_{r=1} w_r^2}^{\text{L2 regularisation}} \right] \qquad \text{Equation 6}$$

The model's source code and data are openly available.[2] Thereby, we aim to support model-building and validation practices. We hope our data can become a benchmark data set for studying choice behaviour in the presence of visual stimuli.

### 3.4.2. Implementation and hyperparameter tuning

Our CV-DCM is implemented and trained in PyTorch (Paszke et al. 2019). PyTorch is a Python-based machine learning package commonly used for deep learning computer vision research because it supports GPU computing. We have determined the hyperparameters of the CV-DCM using a 'heuristic search approach'. We have tried various combinations of optimisation algorithms, learning rates, batch sizes, and regularisation settings to determine the 'optimal' hyperparameters. Ideally, we would have conducted a full-fledged hyperparameter tuning in which all combinations of optimisation algorithms, learning rates, batch sizes, and regularisation parameters would be tested. But, given the computational cost of training CV-DCM (and CV models more generally), we refrained from conducting such a hyperparameter tuning. We found the following hyperparameters to work best (Table 3).

Table 3: Hyperparmeters CV-DCM

| Hyperparameter | Value |
|---|---|
| Optimisation algorithm | Stochastic Gradient Descent |
| Batch size | 20 |
| L2 weight decay ($\gamma$) | 0.1 |
| Learning rate | 1e-6 |

### 3.4.3. Image transformation and feature scaling

In line with common practice in computer vision, we transform and augment images while training the CV-DCM. Specifically, we conduct the two operations. First, we downsampled the images to 224 x 224 pixels. This downsizing operation ensures that images have the input dimensions expected by the CV model (i.e. DeiT base model). Second, we randomly flip images horizontally. This data augmentation operation reduces the model's ability to remember images, thus lowering the chance of overfitting the training data. Furthermore, we scale the numeric features. Scaling the features helps the optimiser to avoid getting stuck in local minima (Géron 2019). The most common type of scaling in machine learning involves shifting and scaling the features to a zero mean and a unit variance. We use another commonly used scaling technique to scale the housing cost and travel time features, called min-max scaling. This scaling entails scaling the features to a range of [-1,1]. The advantage of this scaling technique is that it is straightforward and facilitates easy interpretation of the model's parameters. To facilitate interpretation, we have used the same scaling for all data (thus ignoring that the minimum travel time level varied across respondents, see Table 1). Specifically, all housing costs are divided by 225 and travel times are divided by 15.

### 3.4.4. Train-test split

Splitting the data into a train set and a test set is essential for training virtually all machine learning models because their high capacity makes them prone to overfitting (Géron 2019). As the name suggests, the train set is used for training the model; the test is unseen by the model during training and used to evaluate (test) the model's generalisation performance after training. If a trained model overfits the data, a gap in the performance between the train and test set will tell.

---

[2] github.com/sandervancranenburgh/Computer-vision-enriched-DCMs

The most common way to create the train-test split is by randomly allocating observations to the two sets. When splitting data, it is important to avoid "data leakage". Data leakage happens when the model has access to information during training that it does not have when deployed after training (see e.g. Hillel 2021 for its impact on choice model outcomes). For this study, we split our data across images. Thereby, we aim to avoid potential data leakage from learning the utility levels of specific images rather than generalisable high-level utility-generating features embedded in the images. Making such a split is, however, a nontrivial network problem. Every image is connected at least to one other image (the other street-view image presented in the choice task). However, some images are connected to dozens of other images because they are used more than once (see section 3.3). Hence, when we assign one image to the train set, we must place all directly and *indirectly* connected images in the train data set too.

Given the above 'network' problem, we took the following procedure to create the train and test sets. We randomly picked one choice task, comprising two images, and put this choice task and *all* choice tasks connected to this one in the train set. We repeated the random picking of choice tasks until 80% of the data were used. The remaining data (20%) make the test data set. The train and test data sets comprise, respectively, $N = 9,784$ and $N = 1,948$ choice observations. Due to our splitting strategy, observations of the same individual may be present in both the train and test data sets. However, it is unlikely to cause serious data leakage because no socio-demographic variables (that would be needed to identify observations of the same respondent) are used in the training of the CV-DCM.

## 4. Results
### 4.1. Estimation results

Table 4 shows the main estimation results for the three models, whose utility functions are given in Equation 7 to Equation 9. Models 1 and 2 are standard linear-additive RUM-MNL models used as benchmark models to compare the proposed CV-DCM (Model 3). All models assume decision-makers experience (dis)utility from the housing cost and travel time in a linear and additive fashion. Model 1 ignores the images completely, while Model 2 takes into account the month in which the image is taken by estimating constants, denoted $\beta_{mo}$, for each month. If where and when images are collected are uncorrelated, we expect that images taken in spring and summer, on average, attain a higher utility than images taken in autumn or winter. Model 3 is the proposed CV-DCM and takes the monthly housing cost (*hhc*), commute travel time (*tti*) and the month of the year as numeric input attributes in the same way as Model 2 does, but also takes the feature maps of the images as inputs. For each model, Table 4 shows the performance on the train and test sets, using three (related) metrics: the log-likelihood, rho-square and cross-entropy. A good performance on the test set implies the model generalises well to unseen data. A gap between the performance on the train and test set signals overfitting.

$$U_{in} = \beta_{hhc}hhc_{in} + \beta_{tti}tti_{in} + \varepsilon_{in} \qquad \text{Model 1} \qquad \text{Equation 7}$$

$$U_{in} = \beta_{hhc}hhc_{in} + \beta_{tti}tti_{in} + \sum_{mo}\beta_{mo}I_{S_j} + \varepsilon_{in} \qquad \text{Model 2} \qquad \text{Equation 8}$$

$$U_{in} = \beta_{hhc} hhc_{in} + \beta_{tti} tti_{in} + \sum_{mo} \beta_{mo} I_{S_j} + \sum_{k} \beta_k z_{ikn} + \varepsilon_{in} \quad \text{Model 3} \quad \quad \text{Equation 9}$$

$$\text{where} \quad I_{S_j} := \begin{cases} 1 & \text{if } mo = S_j^{mo} \\ 0 & \text{else} \end{cases},$$

$$z_{ikn} = \phi(S_{in} \mid w_r),$$

$$\varepsilon_{in} \sim \text{i.i.d. Extreme Value Type II}$$

We can draw three conclusions based on the performance metrics in Table 4. The first and most important conclusion is that the CV-DCM can extract relevant information from the street-view images to predict the choice behaviour. Looking at the generalisation performance, we see that CV-DCM outperforms the two benchmark models by a large margin. Specifically, the CV-DCM improves the log-likelihood on the test set by 57 log-likelihood points. The rho-square jumps from 0.116 to 0158. Second, the month of the year carries limited information regarding the utility generated by the images, at least when used in isolation from other information from the images, as in Model 2. Comparing Models 1 and 2, we observe that Model 2 outperforms Model 1 by 23 log-likelihood points on the train set but performs equally well on the test set. Hence, the plain incorporation of the month of the year in the utility function does not improve the generalisability of the standard RUM-MNL models. Third, despite having to train 86 million weights, the extra computational time does not render the CV-DCM impractical; 4.5 hours of training time is in the same order of magnitude as the estimation time of conventional mixed logit models. But, handling large numbers of images and working with GPUs is technically considerably more challenging.

Next, we look at the estimated preference parameters. We see that housing cost and commute travel time are highly relevant attributes to the residential location choice. In line with expectations, $\beta_{tti}$ and $\beta_{hhc}$ are highly significant, and their minus signs align with behavioural intuition. Based on $\beta_{tti}$ and $\beta_{hhc}$, we also compute the VTT.[3] In the context of our SC experiment, the VTT gives the (mean) willingness to pay per month for a one-hour travel time reduction. A VTT between €217 and €228 per hour per month seems reasonable, considering that most respondents in our sample commute five days per week, and thus about 20 days per month. Furthermore, in line with expectations, we observe that the VTT is stable across the three models. We expect stable $\beta_{tti}/\beta_{hhc}$ ratios because our experimental design is constructed such that images and numeric attribute levels within choice tasks are entirely uncorrelated. Cramer (2005) shows that ratios of logit model estimates are unaffected by omitted variables if the omitted variables are uncorrelated with other explanatory variables.

The signs of the estimates associated with the months of the year are mostly intuitive. These estimates reflect the average utility difference between an image taken in that month with images taken in December (which we fixed to zero). In Model 2, we see that the estimates associated with the months of the year are mostly positive and significant for the spring and summer months. This can be explained by the notion that images taken in these months are more likely to look more attractive to live than images taken in winter, for instance, because the weather is better. However, the positive and significant estimate for January counters this line of argumentation and is hard to explain. Now we turn to the estimates associated with the months of the year under Model 3. In Model 3, the estimates associated with the months of the year do not carry the same interpretation as under Model 2. The utility derived from an image in Model 3 is the sum of the utility from the image's feature map and the estimate for the month of the year. As a result, we cannot see the estimates associated with these two utility sources in isolation. One noteworthy observation concerning the estimates related to the months of the year in Model 3 is that fewer estimates are significant than in Model 2. This observation aligns with statistical expectations. Because feature maps already contain information about the month of the year, adding the month of the year explicitly to the model provides little additional information to explain the choice behaviour. To give an example, an image in which trees that have shed their leaves reveals the image is probably taken in winter.

---

[3] $VTT = 60 \left( \frac{225}{15} \right) \frac{\beta_{tti}}{\beta_{hhc}}$ Note that the factor (225/15) comes from the fact that the attributes are scaled before training.

Table 4: Estimation results

|  | | Model 1 | | | Model 2 | | | Model 3 | | |
| --- | --- | --- | --- | --- | --- | --- | --- | --- | --- | --- |
| Model type | | lin-add RUM-MNL | | | lin-add RUM-MNL | | | CV-DCM | | |
| Number of parameters | | 2 | | | 13 | | | 86m | | |
| Estimation time | | <1 sec[I] | | | <1 sec[I] | | | 4.5 hr.[II] | | |
| Train set N = 9784 | Log-Likelihood | -5,954 | | | -5,931 | | | -5,724 | | |
| | $\rho^2$ | 0.120 | | | 0.130 | | | 0.156 | | |
| | Cross-entropy | 0.609 | | | 0.606 | | | 0.585 | | |
| Test set N = 1948 | Log-Likelihood | -1,194 | | | -1,194 | | | -1,137 | | |
| | $\rho^2$ | 0.116 | | | 0.116 | | | 0.158 | | |
| | Cross-entropy | 0.613 | | | 0.613 | | | 0.585 | | |
| | | est | s.e. | p-val | est | s.e. | p-val | est | s.e.[III] | p-val[III] |
| $\beta_{hhc}$ | | -0.86 | 0.025 | 0.00 | -0.87 | 0.024 | 0.00 | -0.96 | 0.025 | 0.00 |
| $\beta_{tti}$ | | -0.21 | 0.023 | 0.00 | -0.21 | 0.025 | 0.00 | -0.24 | 0.026 | 0.00 |
| $\beta_{jan}$ | | | | | 0.46 | 0.129 | 0.00 | 0.25 | 0.136 | 0.07 |
| $\beta_{feb}$ | | | | | 0.02 | 0.228 | 0.91 | -0.40 | 0.240 | 0.10 |
| $\beta_{mar}$ | | | | | 0.10 | 0.080 | 0.23 | 0.05 | 0.084 | 0.58 |
| $\beta_{apr}$ | | | | | 0.25 | 0.080 | 0.00 | 0.36 | 0.084 | 0.00 |
| $\beta_{may}$ | | | | | 0.28 | 0.084 | 0.00 | 0.08 | 0.088 | 0.39 |
| $\beta_{jun}$ | | | | | 0.17 | 0.084 | 0.04 | -0.12 | 0.088 | 0.16 |
| $\beta_{jul}$ | | | | | 0.21 | 0.094 | 0.02 | 0.31 | 0.098 | 0.00 |
| $\beta_{aug}$ | | | | | 0.24 | 0.087 | 0.01 | 0.12 | 0.092 | 0.17 |
| $\beta_{sep}$ | | | | | 0.19 | 0.085 | 0.03 | 0.33 | 0.089 | 0.00 |
| $\beta_{oct}$ | | | | | 0.46 | 0.131 | 0.00 | 0.40 | 0.138 | 0.00 |
| $\beta_{nov}$ | | | | | -0.11 | 0.106 | 0.31 | -0.04 | 0.111 | 0.74 |
| $\beta_{dec}$ | | | | | 0.00 | --fixed | | 0.00 | --fixed | |
| Value-of-Travel-Time [€/hr month] | | 216.7 | 28.26 | 0.00 | 217.2 | 28.35 | 0.00 | 228.5 | 26.73 | 0.00 |

[I] Using 4 CPUs (Xeon @ 3.60 GHz)
[II] Using GPU (GeForce RTX 2080Ti)
[III] Obtained though computing the hessian while keeping the utility derived from the image fixed

Lastly, we analyse the contributions to utility differences between the right and left-hand side alternatives derived from the images' feature maps. To do so, Figure 7 shows three kernel density plots. The left-hand side plot shows the total utility difference as predicted by the trained CV-DCM; the middle plot shows the utility difference from the numeric attributes; and the right-hand side plot shows the utility difference from the images. We make several observations based on Figure 7. Firstly, looking at the range of *x*-axes of the middle and right-hand side plots, we see that the utility differences arising from the images and numeric attributes are similar. This tells us that the part-worth utilities derived from the numeric attributes (housing cost and travel time) are of the same magnitude as those derived from the street-level conditions embedded in the street-view images.[4] This observation adds to the evidence that street-level conditions are important to residential location choice behaviour and can effectively be captured and modelled using images and CV-DCMs. Secondly, we notice that the distributions of utility differences are virtually equal for the test and train sets. This indicates that the CV-DCM does not overfit the training data, and the data are adequately split into train and test sets. Therefore, the CV-DCM must have learned to extract salient generalisable features from the images that generate utility. Thirdly, we see that the distribution of the utility differences stemming from the images is comparatively more bell-shaped than those of the numeric attributes. At first sight, this may seem odd, but it can be explained by how the choice tasks are constructed. Recall that we removed choice tasks without trade-offs between the numeric attributes (see section 3.1.2 for more details). This removal leads to the bi-modal shape of the utility difference.

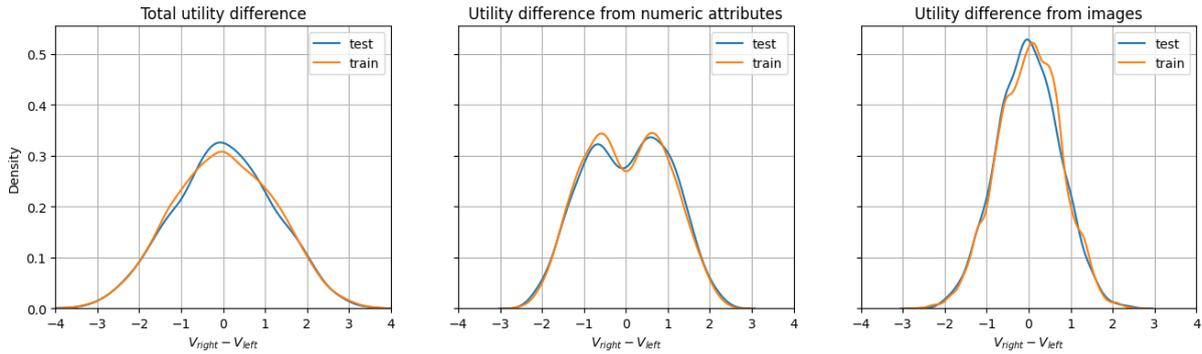

Figure 7: Utility differences

## 4.2. Face validity: what has the CV-DCM learned about street-level conditions?

The considerable performance improvement by the CV-DCM compared to the benchmark models (see Table 4) supports the notion that the CV-DCM can extract relevant information from images to predict choice behaviour. But, $\beta_k$ and $w$ do not carry a behavioural meaning. Therefore, they do not provide directly interpretable insights about what the CV-DCM has learned regarding what decision-makers find important for their residential location choices. To shed light on what the CV-DCM has learned about decision-makers' preferences, we show two collages of images taken from the test set, to which the trained CV-DCM assigns the highest (Figure 8) and lowest (Figure 9) utility levels. Note that the utility level is stamped in the top left of each image.[5] These utility levels are 'uncorrected' for the month of the year. Hence, the top left image yields a utility of 1.63 if the image was taken in December, while it produces a utility of 1.63-0.12 = 1.51 if it was taken in August (which it is).

What catches the eye in Figure 8 is that the images all look spacious, green and often water-abundant. We see many trees, grassland and detached houses. In the authors' view, these street-view images indeed are highly attractive residential locations. In sharp contrast, the images in Figure 9 look cramped, greyish, and urbanised and often have hallmarks of transportation, such as overhead wires, bus stops, parked bikes, and cars. In the authors' view, these street-view images are highly unattractive as residential locations. The mean difference in utility between the 20 best, shown in Figure 8, and the 20 worst images, shown in Figure 9, is 2.7 utility points. The willingness to pay per month to move from the worst to the best street-level conditions can be computed by dividing the utility difference by $\beta_{hhc}$. The result yields a willingness to pay of 632 euros per month – which seems high but not implausible in the authors' view. Here it should be noted that this estimate concerns the two most extreme street-level conditions.

---

[4] Given the ranges of the numeric attributes presented in the SC experiment.
[5] Note that the mean utility derived from images across all images is 0.02 (thus not exactly zero). This is inconsequential as utility has no absolute scale of level, only utility differences matter (Train 2003).

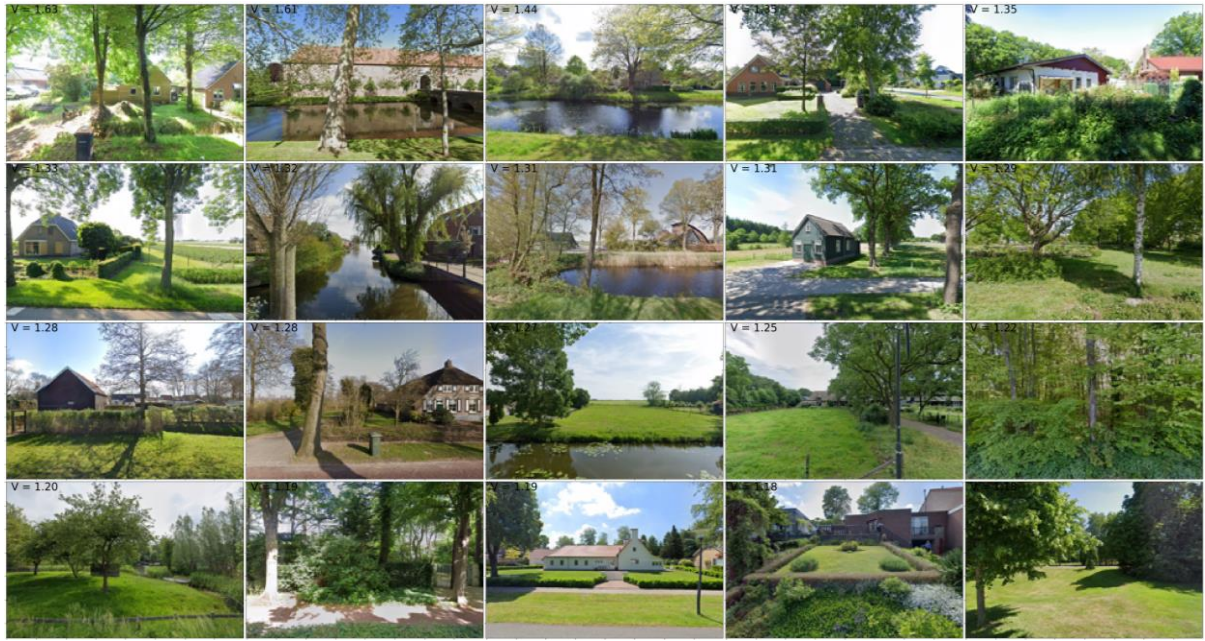

Figure 8: Images with the highest predicted utility levels

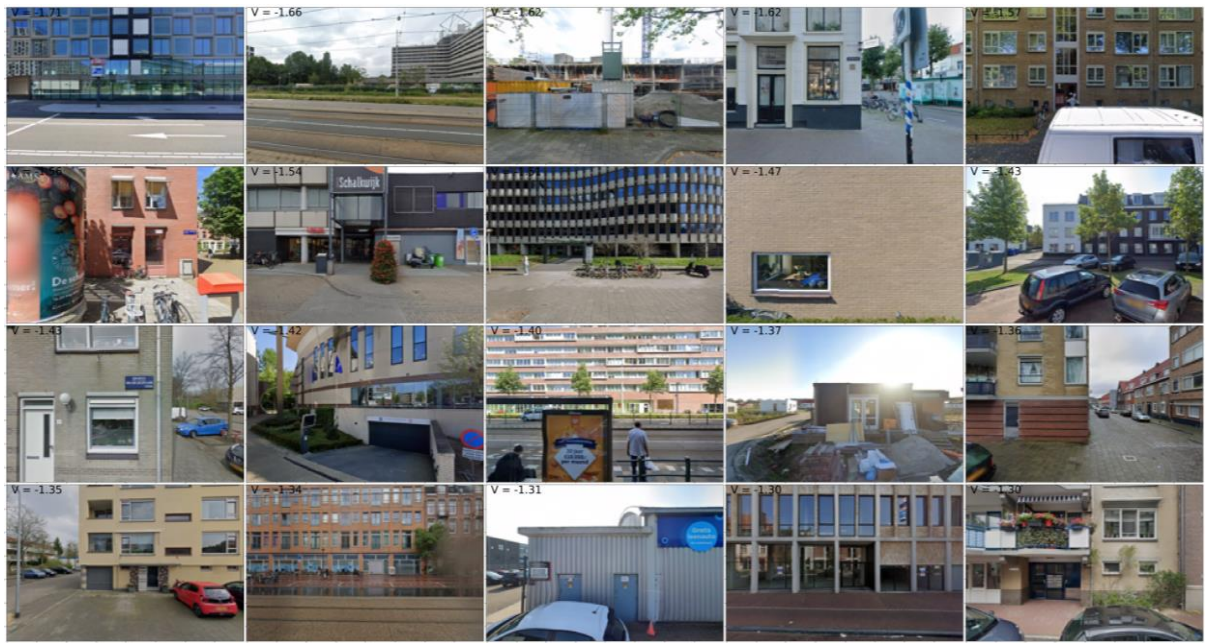

Figure 9: Images with the lowest predicted utility levels

The trained CV-DCM presents an interesting opportunity to investigate the connection between visual attractiveness and population density. Previous research suggests that rural areas are generally considered to be more visually appealing and scenic (Bijker and Haartsen 2012). This heightened visual attractiveness is one of the motivations for counter-urbanism (Elshof et al. 2017), which is characterised by people moving away from urban areas and settling in rural or suburban areas. However, previous studies into counter-urbanism (not using computer vision) relied on proxies for visual attractiveness, such as shares of older housing, proximity to protected natural areas, and a high number of nearby hotels. With our trained CV-DCM, we can examine this relationship more directly. Specifically, for each image in our data set, we calculate the local population density of the 5-level postcode zone it is taken from and group the images based on six population density quantiles. Figure 10 presents the results using a box plot, which shows that increasing population density is associated with lower utility levels from street-level conditions.

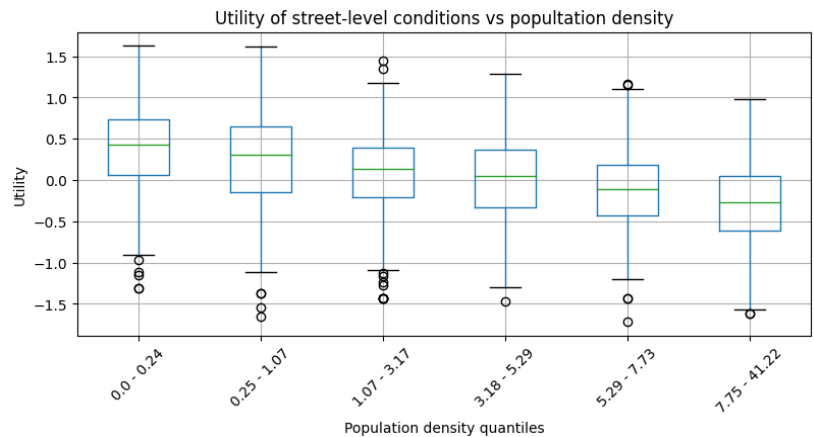

Figure 10: Utility of street-level conditions in high and low-population-density areas

Finally, in order to gain a deeper understanding of the insights gleaned from the CV-DCM, we generated an additional four collages (Figure 11) featuring the highest and lowest utility levels within the highest and lowest population density quantiles. As such, the top-left collage in Figure 11 displays a sample of the most attractive street-view images in low-density areas, while the bottom-right collage displays a sample of the least attractive street-view images in high-density areas. The results of Figure 11 align with our intuition that (un)attractive street-level conditions vary significantly between low and high-density areas. Notably, we observed a marked difference in unattractive street-view conditions between low and high-density areas. In low-density areas, unattractive conditions are primarily caused by views of business parks, while in high-density areas, they are predominantly due to concrete structures and transportation features, such as overhead wires and parked bikes and cars. Consequently, we conclude that the CV-DCM model can recognise the various causes of (un)attractiveness of street-level conditions, which can differ depending on land use.

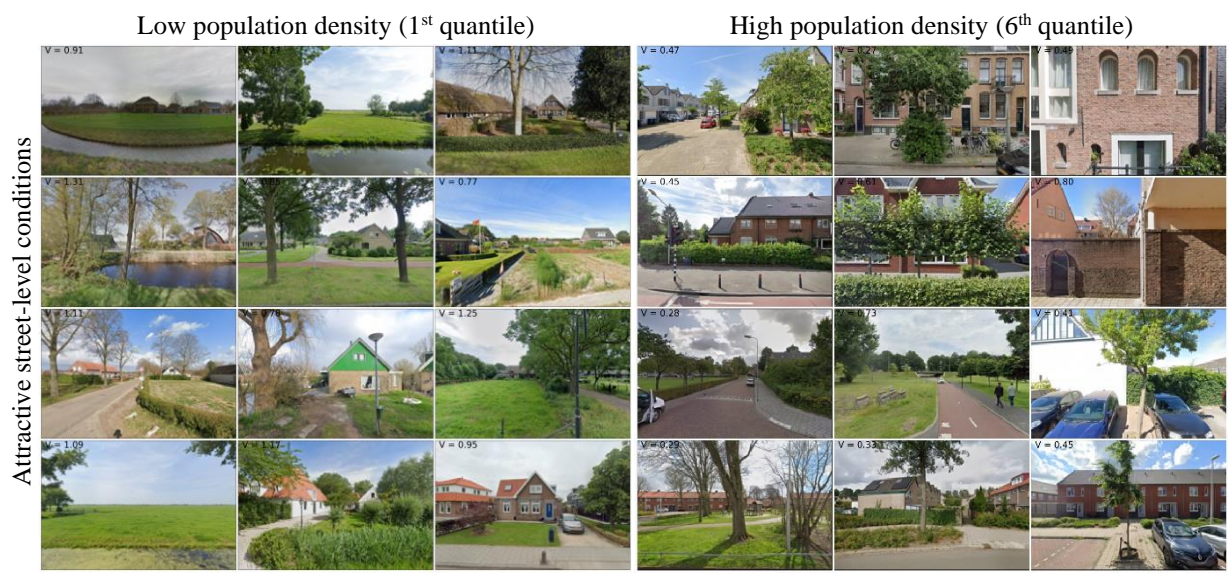

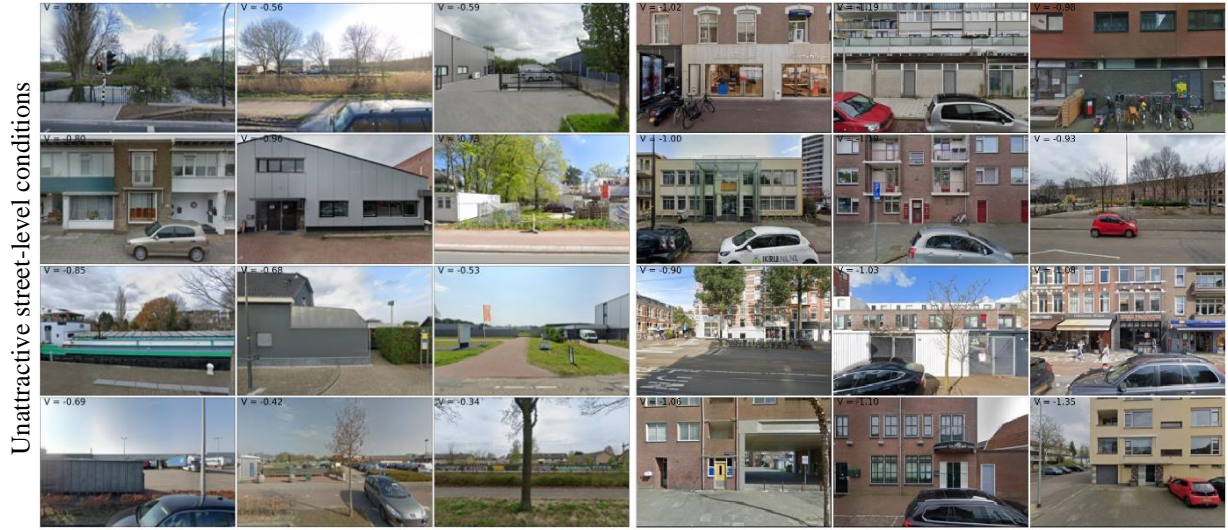

Figure 11: Attractiveness of street-level conditions in high and low population density areas

## 5. Conclusion and discussion

This paper contributes to the recent methodological progress made in the fields of transportation and choice modelling that aims to bring machine learning and DCMs closer together (e.g. Sifringer et al. 2020; Arkoudi et al. 2021; Ramírez et al. 2021; van Cranenburgh et al. 2021). We have proposed a new choice model, called "Computer Vision-enriched Discrete Choice Models", for modelling multi-attribute choice behaviour in the presence of visual and numeric stimuli – methodologically expanding the realm of discrete choice models. The proposed Computer Vision-enriched Discrete Choice Model (abbreviated as CV-DCM) is built from behavioural assumptions, starting with random utility maximisation principles. As such, it has a solid behavioural foundation and can be used to derive marginal utilities and (in principle) willingness to pay estimates. The model should thus be conceived as a behaviour-informed choice model rather than a behaviour-agnostic machine learning model. We have demonstrated its merits by applying it to residential location behaviour –which is strongly coupled with travel demand. We have shown that CV-DCMs can produce new insights into preferences over visual street-level conditions. Notably, we have uncovered which residential places people find most and least attractive and how attractiveness varies with population density.

The proposed model, in conjunction with SC experiments, can potentially enhance the travel behaviour research field's understanding of various other transport-related preferences. Using images can be particularly beneficial when numbers or text fail to convey the choice situation effectively. For instance, preferences related to crowdedness, traffic safety, and spaciousness can be better understood through the use of images in SC experiment showing, e.g. the crowdedness of train platforms, the safety situation of pedestrian crossings, or the extra legroom available when upgrading from economy to business class when booking flights. Incorporating such visualisations can provide valuable insights for transport planners and policymakers seeking to improve transportation systems and services. For instance, inspired by studies like Rossetti et al. (2019), our trained CV-DCM could be deployed to assess the spatial distribution of the attractiveness of large residential areas by feeding all available street-view images and visualising how utility is spatially distributed.

This study raises a plethora of questions and opens up a multitude of avenues for further investigation at the intersection of choice modelling, computer vision and cognitive psychology. Three important questions are of particular interest: (1) how to handle multiple images, (2) how to extract more and better information from trained CV-DCMs and (3) how to effectively employ images and ensure that the features of interest are learned from the images (.e.g as oppose the quality of the image, which would instead reflect the photographer's skills).

Regarding the first question, multiple images are often used to describe alternatives. For instance, real estate websites like Zillow.com and online retailers like Amazon.com often use dozens of images per home or product. While the proposed modelling framework can accommodate a single image per alternative, future research can extend it to enable multiple images (e.g. inspired by Baevski et al., 2022). This methodological advancement would further expand the application domain and enhance the behaviour realism of the discrete choice models.

The second question concerns how to extract more and better information from (trained) CV-DCMs. First, there is reason to believe that more data would help further improve the performance of our CV-DCM and, consequently, its ability to generate refined policy-relevant information. Regarding the question of how to extract information from trained CV-DCMs, the computer vision field can provide good starting points. Recent developments in eXplainable AI (XAI) (Arrieta et al. 2020) offer a range of techniques that can be adapted to extract information from trained CV-DCMs. In particular, they can be leveraged to shed light on what features are learned by the model to explain the choice behaviour and help validate CV-DCMs. Such insights are potentially helpful not only for researchers but also for policymakers and urban planners. For example, in the context of the study's application, XAI techniques should be able to provide insights into the features that make neighbourhoods attractive (as shown in Figure 9) or unattractive (as shown in Figure 10), which can inform the development of policies.

Another area of research that shows promise for extracting additional insights from CV-DCMs is explicitly modelling preference heterogeneity over visual features. Thereby, a more nuanced understanding of the importance of visual features to choice behaviour can be obtained. Different age groups or life stages may appreciate other aspects of the visual environment. For instance, young adults may value vibrancy, while middle-aged people with children may place more importance on street safety. Training the CV-DCM with the DeiT feature extractor in a fully parameterised latent class form is currently technically infeasible (at least for us). One possible way to circumvent this technical barrier is using more lightweight feature extractors, such as TNT-Ti (Han et al. 2021), DeiT-Tiny (Touvron et al. 2021), or Mixer-S/16 (Tolstikhin et al. 2021). Relatedly but more mundane, improving and enriching the utility specification of CV-DCMs seems a promising avenue for further research. For instance, interaction effects between the numeric and visual information, reference effects, and visual correlates of anchoring effects could be modelled (Hevia-Koch and Ladenburg 2019).

The third question pertains to how to effectively use images in SC experiments such that information relevant to researchers can be extracted. The iconic modern art painting "*Ceci n'est pas une pipe*" neatly illustrates this point. The painting depicts a pipe. However, the artist of the painting, René Magritte, claims that it is not a pipe but a painting (readers interested in a more profound discussion of the painting are referred to Foucault 1983). In the context of this study's application, we want to extract information from the image about the attractiveness of the residential location instead of, e.g. which image is most scenic or has the best lighting conditions. Keys to the effective use of images in SC experiments can likely be found in the cognitive psychology field, which is concerned with studying mental processes such as perception, attention, and memory. Their insights can help researchers in our field to understand better how humans perceive and interpret visual information, which, in turn, can guide, e.g. what sort of images to use, how to present images (e.g. in relation to numeric attributes), and how to design SC experiments involving images more generally. For example, cognitive psychologists have long studied the congruity effect (Henik and Tzelgov 1982). The congruity effect describes the tendency for responses to be faster and more accurate when the items being compared are semantically congruent, such as when the word "red" is depicted in red font. In choice experiments, congruity is occasionally used, albeit not well understood. Specifically, in choice experiments, green and red fonts are frequently used to signal a positive or negative direction of an attribute change, with the idea of lowering the cognitive burden for the respondent. Further investigation is needed to determine whether and under what circumstances preferences obtained from discrete choice models are influenced by the medium of information and how notions from cognitive psychology, like congruency, can be leveraged for better preference elicitation.

Finally, to fully harness the complementary information provided by text and images and pursue the avenues for future research outlined above, it is important to note that our modelling tools need a significant push. The current estimation software, survey platforms, computational resources and data handling practices in our field are not geared towards working with (large numbers of) images. Moreover, working with a large number of images generally places higher technical demands on the programming and data-handling skills of researchers. Fortunately, these hurdles are surmountable. Open science practices and actively seeking cross-fertilisation between travel behaviour research, choice modelling, computer vision and cognitive psychology can accelerate progress. By sharing our data and code openly, we hope to contribute to this advancement.

**Acknowledgement**
This work is supported by the TU Delft AI Labs programme.